\documentclass[10pt,twocolumn,letterpaper]{article}

\usepackage[algorithms]{wacv}      
\usepackage{graphicx}
\usepackage{amsmath}
\usepackage{amssymb}
\usepackage{booktabs}
\usepackage[linesnumbered, boxed, ruled]{algorithm2e}
\usepackage{enumitem}
\usepackage{algpseudocode}

\newcommand{\subscript}[2]{$#1 _ #2$}

\usepackage[pagebackref,breaklinks,colorlinks]{hyperref}

\usepackage[capitalize]{cleveref}
\crefname{section}{Sec.}{Secs.}
\Crefname{section}{Section}{Sections}
\Crefname{table}{Table}{Tables}
\crefname{table}{Tab.}{Tabs.}

\def\wacvPaperID{2575} 
\def\confName{WACV}
\def\confYear{2024}

\newcommand{\rulesep}{\unskip\ \vrule\ }
\begin{document}

\title{ Diffusion Brush: Region-Targeted Editing of AI-Generated Images}

\author{Peyman Gholami \qquad Robert Xiao\\
{\tt\small peymang@cs.ubc.ca} \qquad {\tt\small brx@cs.ubc.ca}\\
\small{Computer Science Department, University of British Columbia}
}

\twocolumn[{%
\renewcommand\twocolumn[1][]{#1}%

\maketitle
\setlength{\tabcolsep}{5pt}
\def\qualheight{2.7cm}
\begin{center}
\def\fsh{\small\textbf}
\vspace{-0.5 cm}
\begin{tabular}{ccccc}

                \multicolumn{5}{c}{``masterpiece portrait of a dog, medals, princely, war hero, 8k"}\\
        \frame{\includegraphics[height=\qualheight]{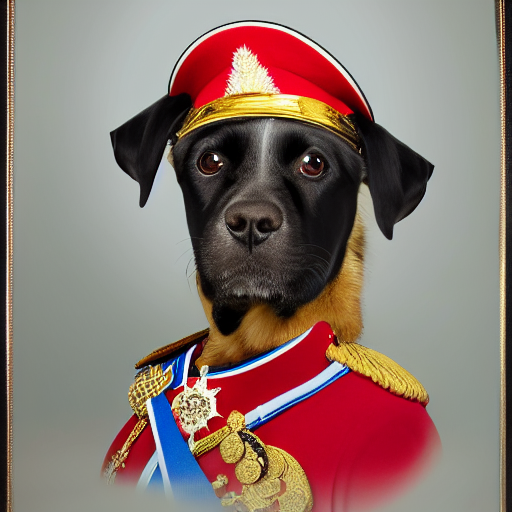}}&
        \frame{\includegraphics[height=\qualheight]{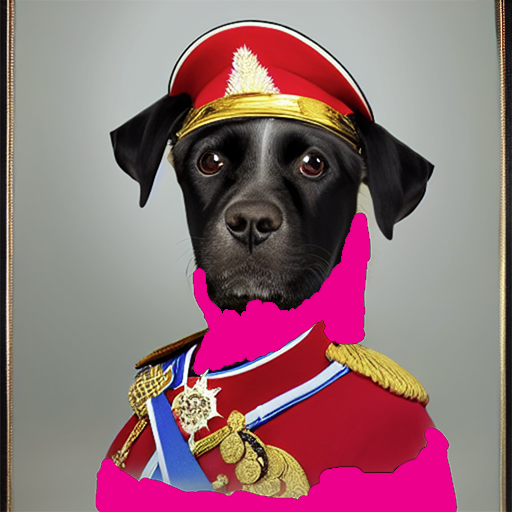}}&
        \frame{\includegraphics[height=\qualheight]{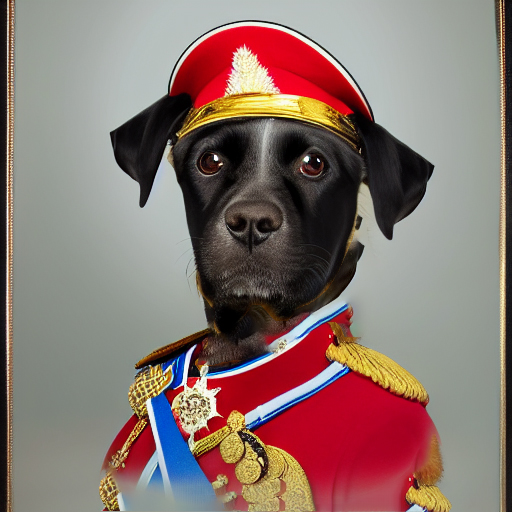}}&
        \frame{\includegraphics[height=\qualheight]{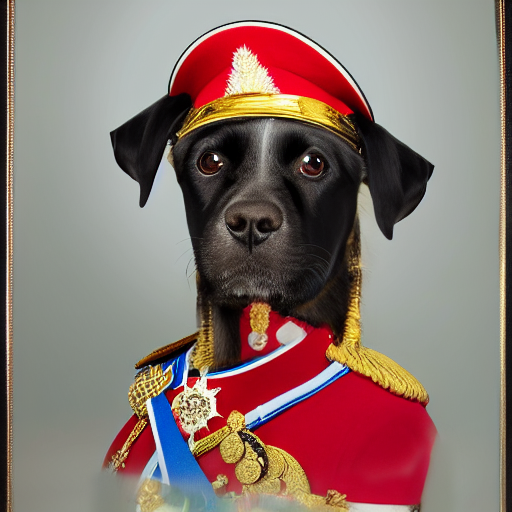}}&
        \frame{\includegraphics[height=\qualheight]{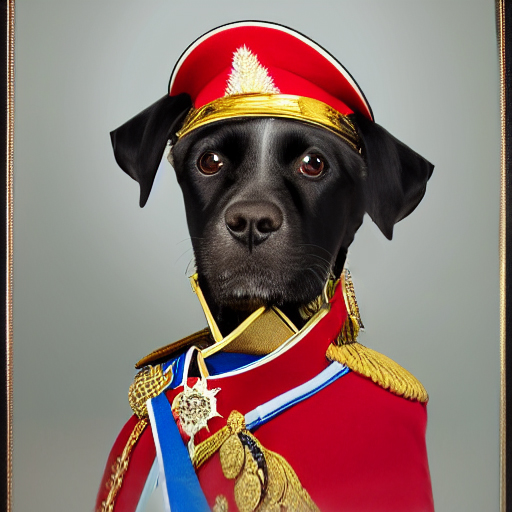}}\\
        \multicolumn{5}{c}{``photo of a monkey eating banana in the city, 4k"}\\
    \frame{\includegraphics[height=\qualheight]{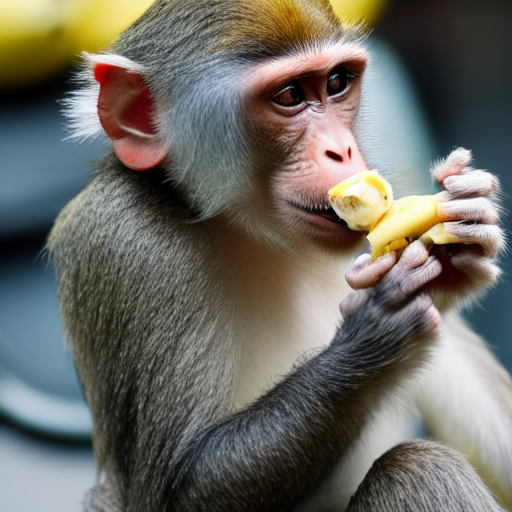}}&
  \label{fig:mon1}
\frame{\includegraphics[height=\qualheight]{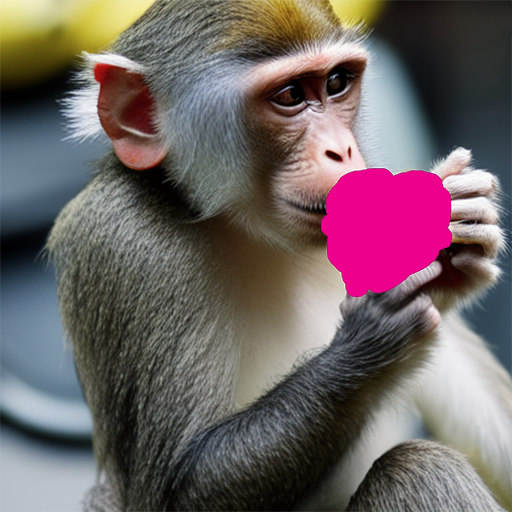}}&
  \label{fig:mon2}
\frame{\includegraphics[height=\qualheight]{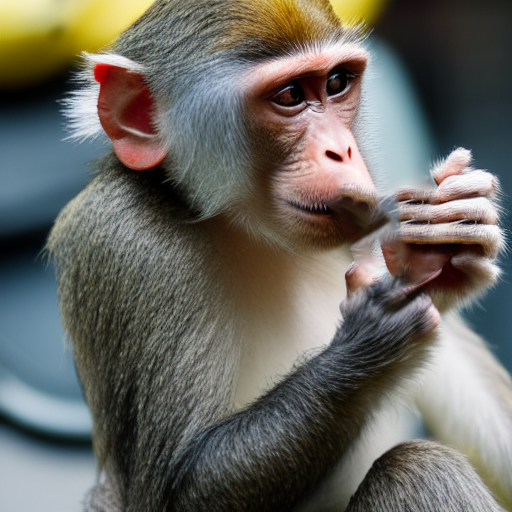}}&
  \label{fig:mon3}
\frame{\includegraphics[height=\qualheight]{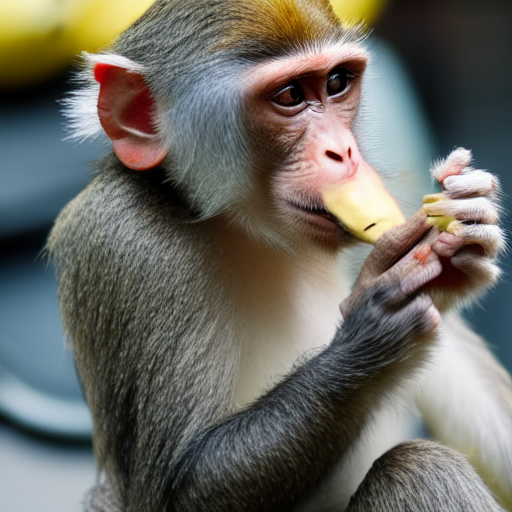}}&
  \label{fig:mon4}
\frame{\includegraphics[height=\qualheight]{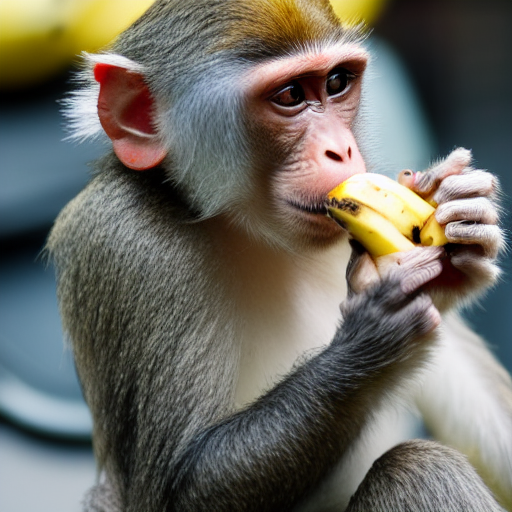}}
  \label{fig:mon5}\\
\multicolumn{5}{c}{``photo of a boat in beach, palm trees, sunset, 4k"}\\
\frame{\includegraphics[height=\qualheight]{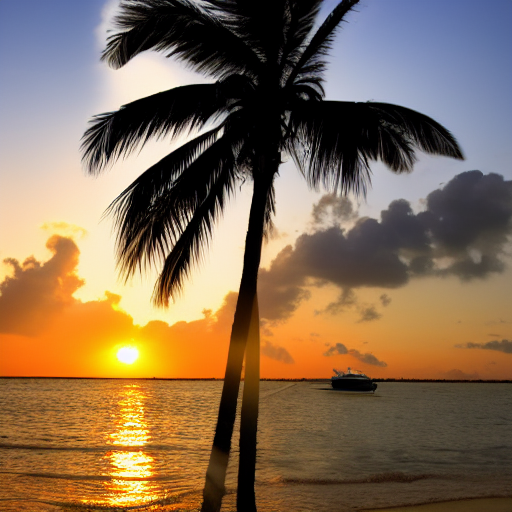}}&
  \label{fig:beach1}
\frame{\includegraphics[height=\qualheight]{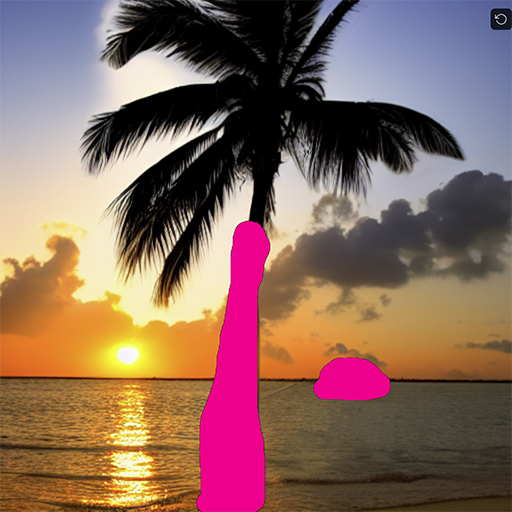}}&
  \label{fig:beach2}
\frame{\includegraphics[height=\qualheight]{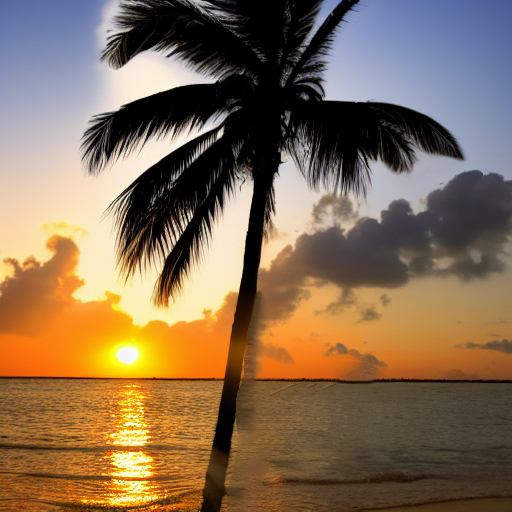}}&
  \label{fig:beach3}
\frame{\includegraphics[height=\qualheight]{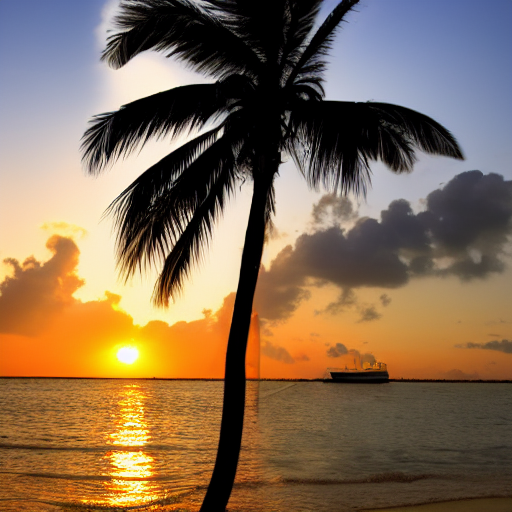}}&
  \label{fig:beach4}
\frame{\includegraphics[height=\qualheight]{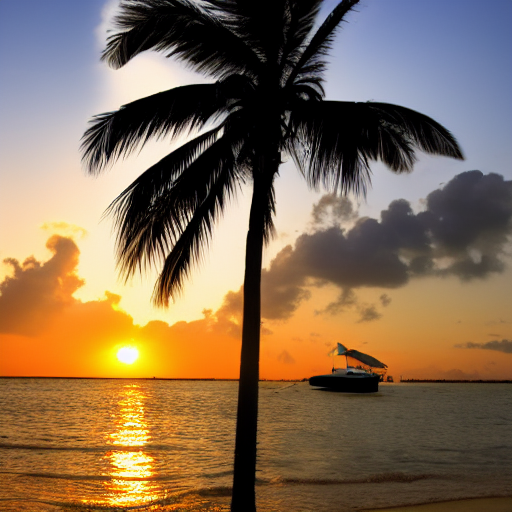}}\\
  \label{fig:beach5}
     \fsh{Original Image} & \fsh{Image with mask} & \fsh{Adobe Photoshop} & \fsh{Inpainting} & \fsh{Diffusion Brush}  \\
\end{tabular}
\captionof{figure}{Comparison of different fine-tuning methods for sample generated images. We present the fine-tuning results for Adobe Photoshop's content-aware filling feature, Inpainting \cite{rombach2022high} method with the same prompt, and Diffusion Brush.}
\label{fig:sample_inpainting}
\end{center}
}]


\begin{abstract}

  Text-to-image generative models have made remarkable advancements in generating high-quality images. However, generated images often contain undesirable artifacts or other errors due to model limitations. Common techniques to fine-tune generated images are time-consuming (manual editing), produce poorly-integrated results (inpainting), or result in unexpected changes across the entire image (variation selection and prompt fine-tuning). In this work, we present Diffusion Brush, a Latent Diffusion Model-based (LDM) tool to efficiently fine-tune desired regions within an AI-synthesized image. Our method introduces new random noise patterns at targeted regions during the reverse diffusion process, enabling the model to make changes to the specified regions while preserving the original context for the rest of the image. We evaluate our method's usability and effectiveness through a user study with artists, comparing our technique against other common image inpainting techniques and editing software for fine-tuning AI-generated imagery.
\end{abstract}


\section{Introduction} \label{sec:intro}
Text-to-image generative models are a class of machine learning models that generate images guided by textual descriptions, and include techniques such as Generative Adversarial Networks (GANs), Variational Autoencoders (VAEs), and Diffusion-based Models (DMs) \cite{zhang2023text}. DMs, which have recently exploded in popularity, have garnered considerable attention due to various characteristics, including a simple training scheme, great performance in generating high-quality images with a wide range of styles, and not being prone to mode collapse. DMs are trained by adding noise to an image. The model is trained to eliminate the noise and restore the original image, for varying levels of noise. To generate images, an initial noise pattern is generated and progressively ``denoised'' using the trained model. Models can be conditioned with text (trained using image-caption pairs), allowing them to generate images based on textual prompts. Latent Diffusion Models (LDMs) are a variant of DMs that use pre-trained autoencoders to shift the diffusion process to a latent space \cite{rombach2022high}. By doing this, models can operate on smaller latent-space images, making them more computationally efficient without losing quality. \cite{ho2022cascaded}.

Despite showing exceptional results in different applications, e.g. image synthesis, super-resolution \cite{saharia2022image}, inpainting, etc., these models are prone to several shortcomings, such as being sensitive to the choice of the original noise distribution and being difficult to control \cite{wang2020state}. These limitations can lead to a high degree of stochastic behavior, often requiring many ``generations'' before a desirable result is achieved. Additionally, due to the nature of the LDM generation technique, the results are often hard to fine-tune. The majority of existing tools for fine-tuning generated images are hard to target, generating global changes when only local changes are desired, or producing alterations that are poorly integrated with the original image. On the other hand, LDM models commonly produce some errors and flaws over different regions of the generated image, that are hard to fix using the existing methods.

In this work, we introduce Diffusion Brush, a novel LDM-based editing and fine-tuning tool that users, including artists, can utilize for making efficiently targeted adjustments and modifications to AI-generated images. Diffusion Brush presents the user with a set of stacked masks for editing purposes, which are used to mask out areas of the image that will be regenerated by the model. Additionally, our tool features a user-friendly interface and a controllable set of parameters, allowing users to quickly repair multiple defects in generated images.

We integrate our method with Stable Diffusion \cite{rombach2022high} and incorporate the ability to create and merge masks with varying strengths at different diffusion steps. Our method introduces new random noise patterns at targeted regions in the reverse diffusion process and combines them with the original image latent intermediately, enabling the model to make targeted adjustments to the specified regions while preserving the original context from the rest of the image, thus allowing the changed regions to be well-integrated with the rest of the generated image without causing global changes.

To evaluate the performance, usability, and effectiveness of our tool, we conducted a user study with artists who are experienced in editing images and also have basic familiarity with AI image editing and generative tools. Our user study aimed to assess the efficacy of Diffusion Brush in improving the experience of users in fine-tuning AI-generated images and providing artistic control in comparison to other existing image editing tools.

\section{Related Work}

Image editing is a continuously evolving area of research in image processing, which has seen significant advancements in recent years. One such development has been the increasing utilization of GANs in a range of image manipulation tasks \cite{abdal2021styleflow, lang2021explaining, pan2023drag}. However, one of the main challenges faced in using GANs for image editing is the process of GAN-inversion \cite{richardson2021encoding}. While producing decent global changes, these methods often fail to create localized edits\cite{bar2022text2live}. Recently, DragGAN \cite{pan2023drag} proposed a GAN-based image editing for the controlled deformation of objects in an image. Their method shows interesting results for the manipulation of the pose and layout of certain objects, but it lacks generalizability and is limited to certain scenarios.

As discussed in section \ref{sec:intro}, LDMs offer a different approach to image editing by learning the underlying distribution of image data and generating new images based on that. As such, many studies have attempted to harness the power of controllable DMs in various image editing tasks. Text and image-driven image manipulation studies are examples of controllable generative models\cite{kim2022diffusionclip,hu2022global}.
Many such approaches, while attempting to make localized changes to images end up creating some degree of global change as well, which is not desirable. Some recent techniques utilize inversion \cite{choi2021ilvr} in order to preserve a subject while modifying the context. Textual Inversion \cite{gal2022image} and DreamBooth \cite{ruiz2023dreambooth} synthesize new views of a given subject given 3–5 images of the subject and a target text.
Several other works \cite{bar2022text2live, brooks2022instructpix2pix, kawar2023imagic, couairon2022diffedit, hertz2022prompt} have addressed the problem of text-based image manipulation and editing.
These methods provide a good basis for basic and describable or subject-related modifications to an image, however, they are not efficient for making fine-tuning modifications that cannot be described with a single prompt. 

Another common image manipulation category is image inpainting, i.e., the task of adding new content to an image. Traditional inpainting can also be used for object removal by not providing any text prompt as guidance \cite{avrahami2022blended}. Even though these models are effective for new content generation in images, they are not appropriate for making small and targeted adjustments \cite{avrahami2022blended, lugmayr2022repaint, saharia2022palette}. However, in contrast, our goal is to fine-tune the image and make localized adjustments to the existing image, while keeping the remaining content of the image intact. Inpainting typically involves using a user-specified mask to provide region-based editing. In this work, we will use a similar masking strategy to control the diffusion process. We also compare our proposed method with the stable diffusion inpainting \cite{rombach2022high}.

\section{Methods}
\subsection{Diffusion Brush Formulation}
As discussed in section \ref{sec:intro}, we use an LDM-based variant of image generative models. Similar to \cite{lugmayr2022repaint}, we make intermediate adjustments to the latent space using the pre-trained LDM. Therefore, Diffusion Brush does not require any additional training and only modifies the latents during the reverse diffusion process.
\subsubsection{LDM Formulation}
DMs solve the problem of image generation by randomly sampling a noise image $x_0 \sim \mathcal{N}(0, \sigma^2_{max}I)$, and sequentially denoises it into images $x_i$ where the noise level of $\sigma_0 = \sigma_{max} > \sigma_1 > \cdots > \sigma_N = 0$. The sequential denoising can be formulated through the simulation of an ordinary differential equation (ODE) term and a
stochastic differential equation (SDE) term as follows:
\cite{karras2022elucidating}:  

\begin{equation}
\begin{split}
    x=\underbrace{-\dot{\sigma}(t) \sigma(t) \nabla_{\mathbf{x}} \log p(\mathbf{x} ; \sigma(t)) d t}_{\text {Probability Flow ODE}}-\\
    \underbrace{\beta(t) \sigma^2(t) \nabla_{\mathbf{x}} \log p(\mathbf{x} ; \sigma(t)) d t+\sqrt{2 \beta(t)} \sigma(t) d \omega_t}_{\text {Langevin diffusion component }},
    \end{split}
\end{equation}
Where $x$ is the sample, $p(\mathbf{x};\sigma)$ is obtained by adding i.i.d. Gaussian noise of standard deviation $\sigma$ to the data, $\omega_t$ is the standard Wiener process, and $\nabla_{\mathbf{x}} \log p(\mathbf{x} ; \sigma(t))$ is ], a vector field that
points towards the higher density of data at a given noise level. We use the formulation of the Stochastic Sampler algorithm with the Euler steps method \cite{karras2022elucidating} for solving the above equation. The original pseudo-code for this algorithm can be found in \cite{karras2022elucidating}. 

We modify the algorithm as follows. The user first creates any number of editing masks $m_1-m_M$. For each mask $m_k$, the user specifies an intermediate step number $n_k$ (per \cref{fig:overview}), which is the steps at which the noise is introduced to the latent, and a tunable parameter $\alpha_k$ to control the additive noise strength.

We then initialize the sample $x_0 = \epsilon_0$ and noise level $\sigma_0$ ($i = 0)$. At each step, for a total number of $N$ steps, we sample noise from $\mathcal{N}(0, S^2_{noise}I)$. Then, at step $n$,  we generate a new noise pattern $x^{\prime}_k  = \epsilon^{\prime}_k$ sampled from  $\mathcal{N}(0, S^{\prime2}_{noise}I)$. Here we are generating a new noise pattern from a different seed. The new noise is added to the original latent controlled by the mask and $\alpha_k$. We will discuss the effect of using different values for $n_k$ in \cref{ui} in detail. 

Finally, at step $t$, the latent of the new seed and the original seed will be merged using the mask, thus bounding the influence of the new latent to the masked region. Subsequently, the new latent region will be progressively denoised and integrated into the existing image contents in steps $t$ through $N$. To keep the UI simple and reduce the number of user-controlled parameters, we fix the value of $t$ in our implementation. Preliminary tests suggested the value of $t = N - 10$ works well for many applications. If multiple masks are used, the process of generation after $x^{\prime}_{n+1}$ is repeated for each mask.

We show the overview of the core algorithm of the proposed method in \cref{fig:overview} and also present the updated pseudocode for the denoising process:

\begin{figure*}[!ht]
    \centering
    \includegraphics[width=\textwidth]{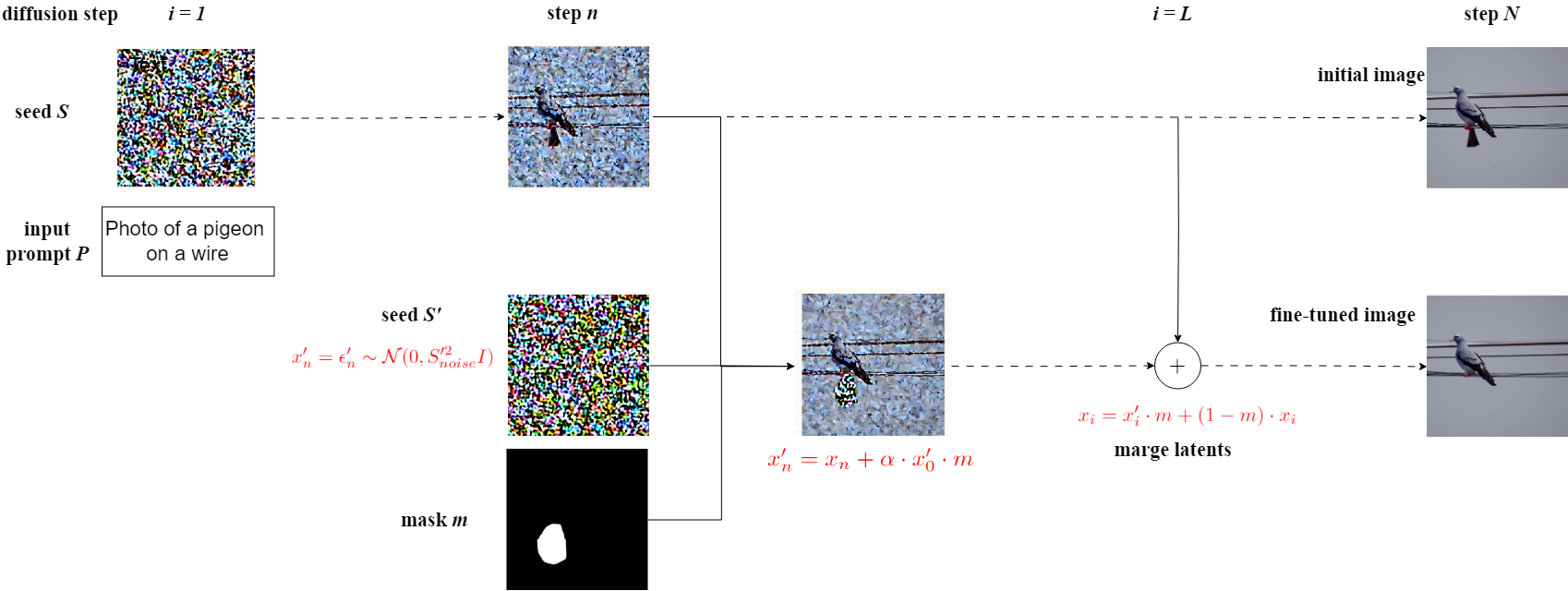}
    \caption{Overview of the proposed method: The top row shows the typical DM generation of an image using a seed $S$ and a prompt $P$. The bottom shows the overview of the proposed method: the algorithm takes a new seed $S'$, and combines it with the original latent at step $n$ using a mask $m$, and a strength control $alpha$. The diffusion process moves forward and at a certain step $t$ (we fix $t$ = $N-10$) the original latent and the new modified latent are merged together using the masking and finally the fine-tuned image is generated. }
    \label{fig:overview}
\end{figure*}

\begin{algorithm}\label{alg:diff_algo}

\SetAlgoLined
\textbf{Input:} A source prompt \textbf{$\mathcal{P}$}, and a random seed \textbf{$S$} for generating original image; A fine-tuning mask \textbf{$m$}, a random seed \textbf{$S'$}, a mask strength control variable \textbf{$\alpha$}, a step number \textbf{$n$} to introduce additive noise\\
\textbf{Output:} Fine-tuned image $x_{N}$.\\
 $x_0 \sim \mathcal{N} (0,S^2 I)$ \Comment{Generate initial sample} \\ 
 \For{$i=0,1,\ldots,N$}{
    \Comment{i = number of diffusion step}\\
    \For{$k=0,1,\ldots,M$}{
        \Comment{k = index of the mask}\\
        $x_{i+1}  \gets DM(x^{\prime}_i , P, i, S)$\\
     \uIf{$i==n_k$}{
    $x^{\prime}_0  \gets \epsilon^{\prime}_{n_k} \sim \mathcal{N}(0, S^{\prime2}I)$\\ 
    $x^{\prime}_{n_k} \gets x_{n_k} + \alpha \cdot x^{\prime}_0 \cdot m_k$\Comment{add the new noise}\\
  }
  \ElseIf {$i>=n_k$}{
    $x^{\prime}_{i+1}  \gets DM(x^{\prime}_i , P, i, S^{\prime})$\\
  }
  \uIf{i==t}{
   $x_{i} \gets x^{\prime}_{i} \cdot m_k + (1 - m_k) \cdot x_{i}$ \Comment{merge the latents}\\
  }
  }
 }
 \textbf{Return} $x_N$
\caption{Diffusion Brush image fine-tuning}
\end{algorithm}

\subsection{User-Interface Design}
\label{ui}

\begin{figure*}[!ht]
  \centering
      \begin{minipage}[b]{0.16\textwidth}
    \includegraphics[width=\textwidth]{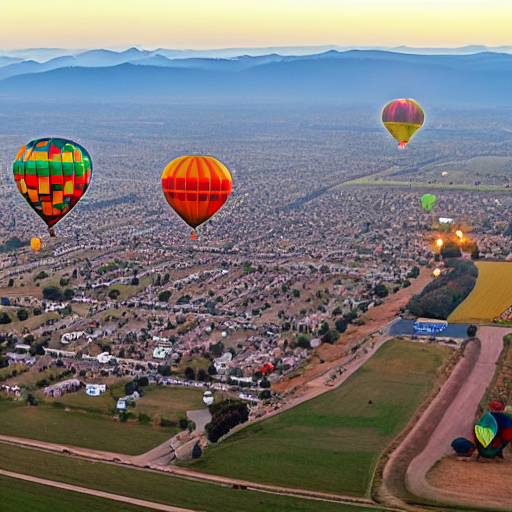}
        \caption*{\small original image}
  \end{minipage} \rulesep
    \begin{minipage}[b]{0.16\textwidth}
    \includegraphics[width=\textwidth]{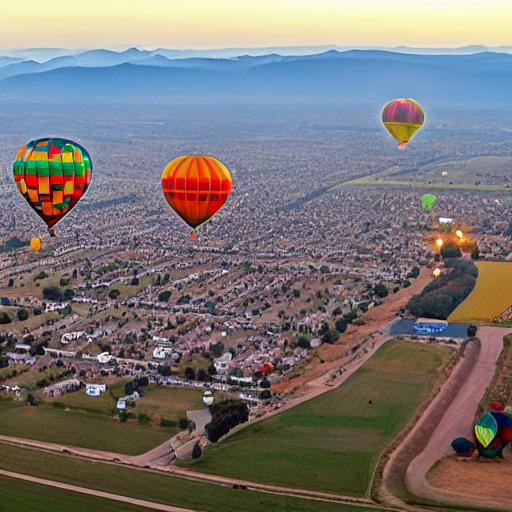}
        \caption*{$\alpha = 0.1$, $n=10$}
  \end{minipage}
    \begin{minipage}[b]{0.16\textwidth}
    \includegraphics[width=\textwidth]{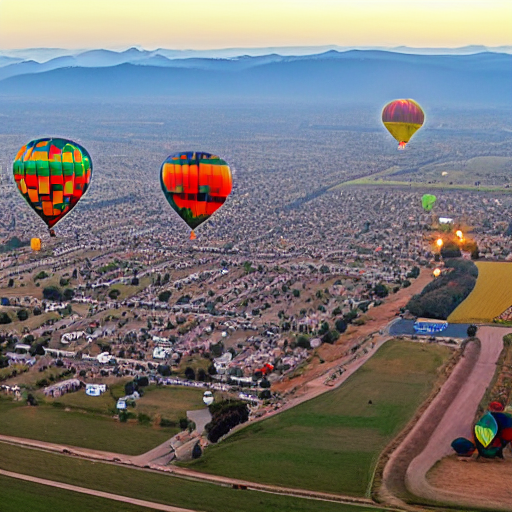}
        \caption*{$\alpha = 1.3$, $n=10$}
  \end{minipage}
      \begin{minipage}[b]{0.16\textwidth}
    \includegraphics[width=\textwidth]{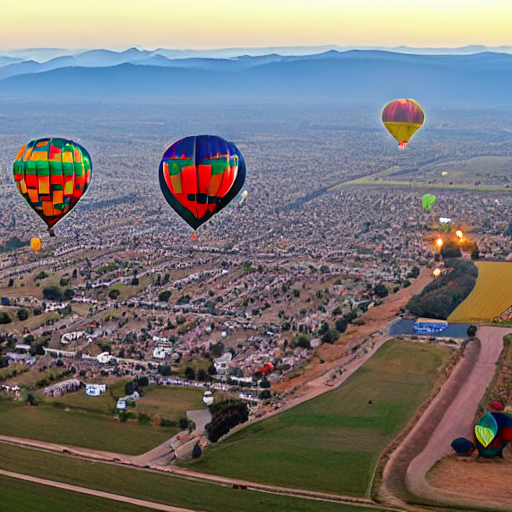}
        \caption*{$\alpha = 2$, $n=10$}
  \end{minipage}
      \begin{minipage}[b]{0.16\textwidth}
    \includegraphics[width=\textwidth]{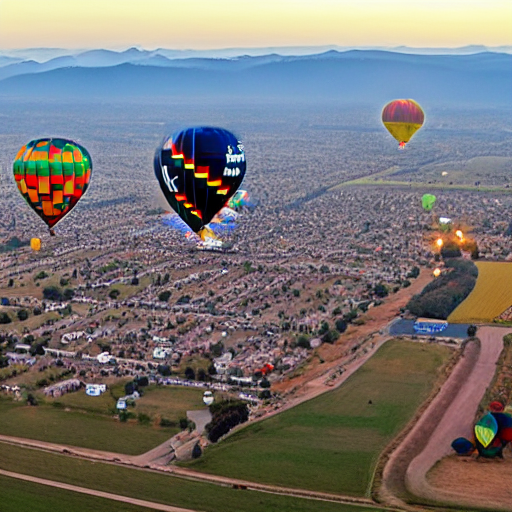}
        \caption*{$\alpha = 4$, $n=10$}
  \end{minipage}
        \begin{minipage}[b]{0.16\textwidth}
    \includegraphics[width=\textwidth]{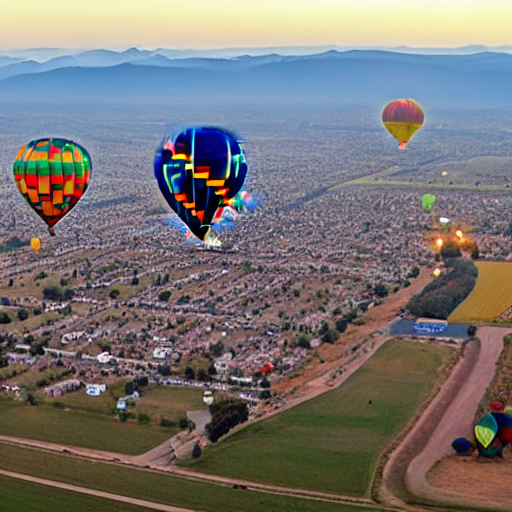}
        \caption*{$\alpha = 5$, $n=10$}
  \end{minipage}
        \begin{minipage}[b]{0.16\textwidth}
    \includegraphics[width=\textwidth]{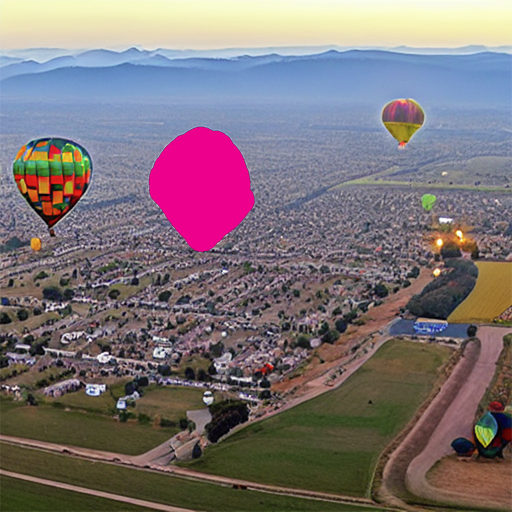}
        \caption*{\small image with mask}
  \end{minipage}\rulesep
    \begin{minipage}[b]{0.16\textwidth}
    \includegraphics[width=\textwidth]{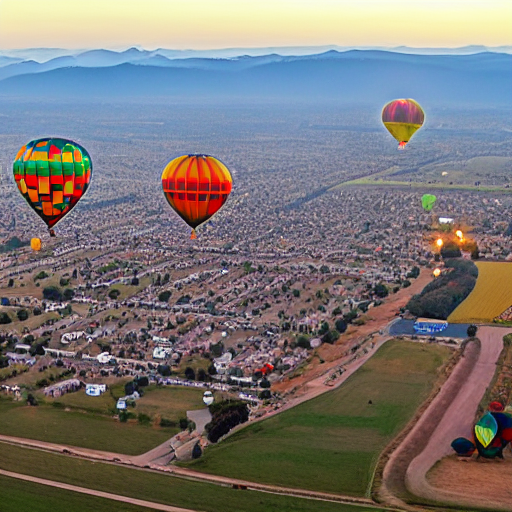}
        \caption*{$\alpha = 1$, $n=0$}
  \end{minipage}
    \begin{minipage}[b]{0.16\textwidth}
    \includegraphics[width=\textwidth]{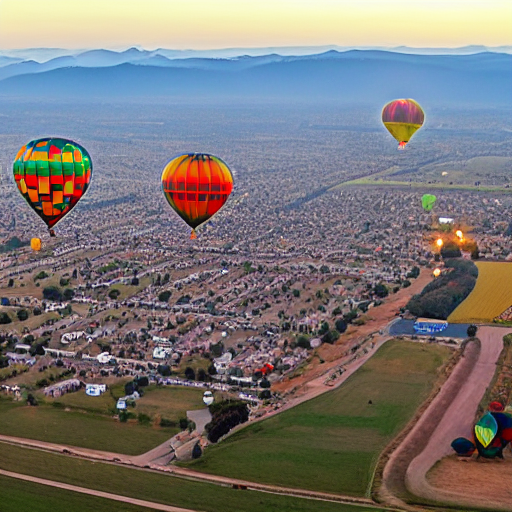}
        \caption*{$\alpha = 1$, $n=5$}
  \end{minipage}
      \begin{minipage}[b]{0.16\textwidth}
    \includegraphics[width=\textwidth]{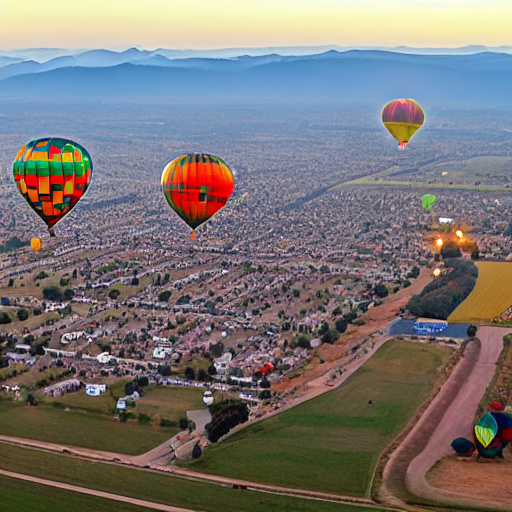}
        \caption*{$\alpha = 1$, $n=15$}
  \end{minipage}
      \begin{minipage}[b]{0.16\textwidth}
        \includegraphics[width=\textwidth]{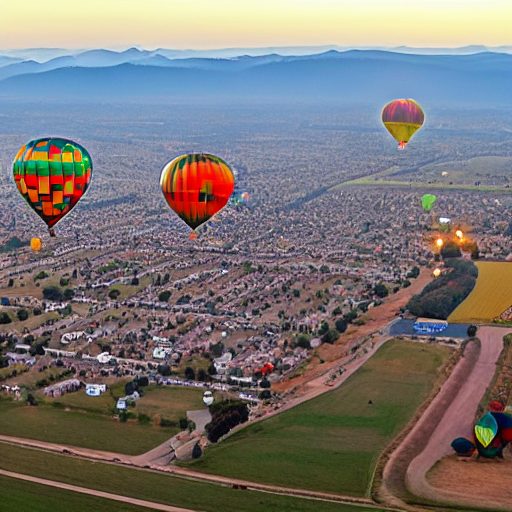}
        \caption*{$\alpha = 1$, $n=25$}
  \end{minipage}
        \begin{minipage}[b]{0.16\textwidth}
    \includegraphics[width=\textwidth]{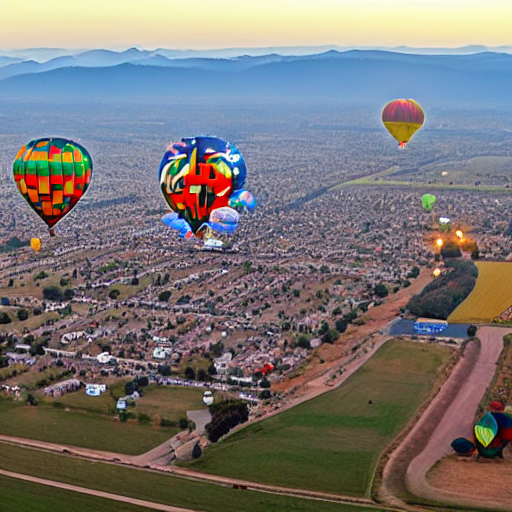}
        \caption*{$\alpha = 1$, $n=35$}
  \end{minipage}
\caption{Effect of changing different controlling parameters of the Diffusion Brush. On the top row, the mask strength ($\alpha$) is altered while the mask, the seed, and the intermediate number for denoising ($n$) have remained stationary. On the bottom row, we changed $n$  and the rest of the parameters remained intact. In all examples $s = 3485530643$ and the total number of diffusion steps is 50}
\label{stren}
\end{figure*}

We build our UI on top of the \href{https://github.com/AUTOMATIC1111/stable-diffusion-webui}{Stable Diffusion WebUI} \cite{SDWUI}, a popular browser-based front-end to the Stable Diffusion series of LDM models. Diffusion Brush appears as a panel where users can create various masks on top of a loaded or previously-generated image, configure the parameters per-mask, and enable/disable them using the checkboxes. The UI provides controls for the following hyperparameters per mask:
\begin{itemize}

\item \textbf{Step Number ($n$)}: An integer that specifies the intermediate step at which the new noise pattern is added to the original image latent space.
\item \textbf{Mask Strength ($\alpha$)}: A number that corresponds to the $\alpha$ value which controls how strong the noise should be applied at the targeted region at step $n$.
\item \textbf{Seed Number ($S$)}: An integer that will be used for generating the Gaussian noise pattern in the specified region. If $S=-1$, a random noise pattern will be used. As with normal image generation, the UI provides buttons to randomize the seed or reuse the previous seed.

\end{itemize}
The step number and the mask strength parameters together control the magnitude of the intermediate additive noise and ultimately the amount of applied change to the latent image.
Higher values for the mask strength will result in a bigger change. Figure \ref{stren} top row shows the effect of a wide range of different mask strength values while keeping other parameters the same. As shown, if the mask strength is too high, the LDM cannot recover, and the additive noise causes artifacts. Conversely, if the $\alpha$ value is too small, Diffusion Brush will not be able to make sufficient adjustments. Similarly, we tested the effect of using different $n$ values while keeping the rest of the parameters stationary. As demonstrated, if the new noise pattern is introduced in the final steps, the LDM cannot recover from it and will produce artifacts.

Additionally, there is a correlation between the two parameters. When the noise is introduced in later stages of diffusion (higher value of $n$), the model has less time to recover from the additive noise. Therefore, it is recommended to adjust the $alpha$ value to a smaller number. Conversely, if the noise is introduced in the early stages of diffusion, and if the value of $alpha$ is not large enough, the additive noise will be dissolved into the original image latent and several stages of diffusion will eliminate its effect. Therefore, if $n$ is small, the $alpha$ value should be larger to achieve optimal results.

\begin{figure*}[!ht]
  \centering
  \begin{minipage}[b]{0.24\textwidth}
        \includegraphics[width=\textwidth]{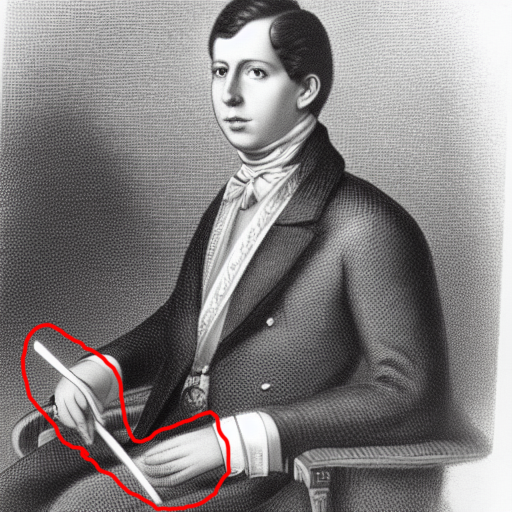}
  \end{minipage}
    \begin{minipage}[b]{0.24\textwidth}
        \includegraphics[width=\textwidth]{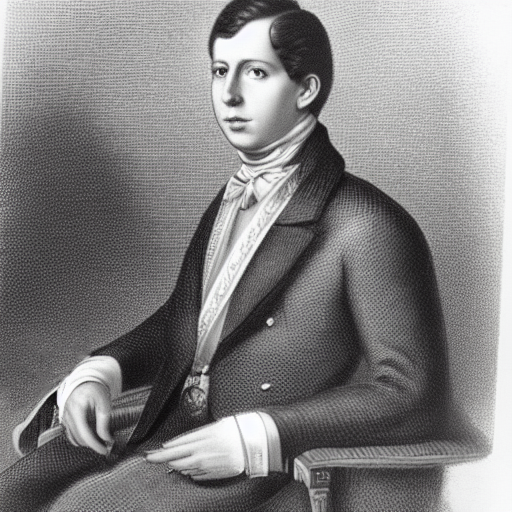}
  \end{minipage}
    \begin{minipage}[b]{0.24\textwidth}
        \includegraphics[width=\textwidth]{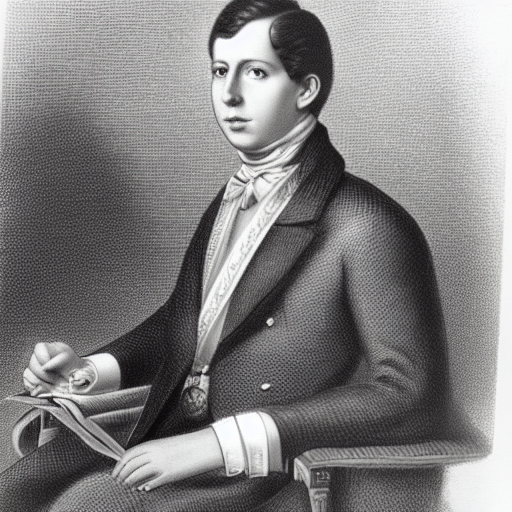}
  \end{minipage}
    \begin{minipage}[b]{0.24\textwidth}
        \includegraphics[width=\textwidth]{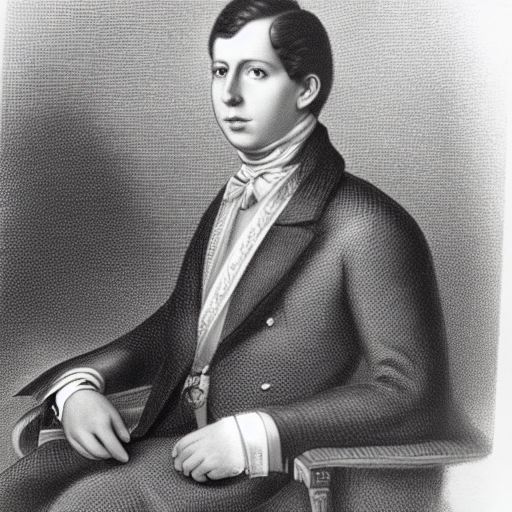}
  \end{minipage}

\caption{Effect of using different seed numbers with the same set of parameters (i,e, the mask, $n$, and $\alpha$).}
  \label{seedimg}
\end{figure*}

Once the masks are specified and the parameters are set, the image generation can be triggered. During the diffusion process, intermediate latent space outputs for the original image and the new seed are previewed to provide visual feedback to the user. This allows the user to make further adjustments in the next steps if they decide to keep the current seed.

In Figure \ref{seedimg} we showcase the effect of using different seeds with the same set of parameters. As demonstrated, Diffusion Brush can be used with different seeds and the same parameters to generate various fine-tuned images. This allows users to select the image with the desired changes, making the tool versatile and customizable.


\subsection{User Study}
\label{user-study}
Due to the nature of the problem that this study tries to address, We conduct a user study in order to evaluate the effectiveness of the Diffusion Brush for providing targeted image fine-tuning in comparison with Inpainting and Adobe Photoshop content-aware fill followed by manual editing (optional).

\subsubsection{Participants}

In this research, we enrolled a group of five expert participants with varying levels of expertise in AI image generation techniques and Adobe Photoshop. Our cohort consisted of two females and three males, with an average age of 27.6 years. As part of our participant selection criteria, we made sure that all of them had at least a basic level of familiarity with AI image generation techniques and were also frequent users of editing software such as Adobe Photoshop for creating visual art. Two participants were well-versed in image generative models and Stable Diffusion, while the remaining three were daily users of Photoshop. 

\subsubsection{Study Procedure and Task Description}
The study began with a brief introduction to the LDM image generation techniques and the diffusion process. Following the introduction, the participants were given a tutorial on how to use the Stable Diffusion WebUI. They were then given a 5-minute window to explore different parameters and the process of generating a new image using a random seed and a prompt.

Each participant was provided with five previously generated images that contained some level of imperfection. They were then asked to fix these images using three different tools: Diffusion Brush, Inpainting, and Adobe Photoshop's content-aware fill tool. The participants were given a maximum of three minutes to edit each image using each of the three tools. The completion times for each task were recorded for each participant.

\subsubsection{Evaluation Survey}
After completing the image repair tasks, the participants were asked to complete a System Usability Scale (SUS) form to rate the usability, ease of use, design, and performance of each method. SUS is a standard usability evaluation survey which is widely used in user-experience literature \cite{brooke1996sus}. 
The participants were presented with 10 questions about each of the methods and were asked to rate each system on a scale of 1 to 5 for each question. A rating of 1 indicated strong disagreement, while a rating of 5 indicated strong agreement. The questions were designed to assess the participants' perceptions of the effectiveness, ease of use, and overall user experience of each tool. Below is the list of the questions:

\begin{enumerate}[label=\subscript{\textbf{Q}}{\textbf{{\arabic*}}}]
\itemsep0em 
    \item I think that I would like to use this tool frequently.
    \item I found the tool unnecessarily complex.
    \item I thought the tool was easy to use.
    \item I think that I would need the support of a technical person to be able to use this tool.
    \item I found the various functions in this tool were well integrated.
    \item I thought there was too much inconsistency in this tool.
    \item I would imagine that most people would learn to use this tool very quickly.
    \item I found the tool very cumbersome to use.
    \item I felt very confident using the tool.
    \item I needed to learn a lot of things before I could get going with this tool.

\end{enumerate}

SUS consists of positive and negative phrasing questions. Q2, 4, 6, 8, and 10 are negatively framed and Q1, 3, 5, 7, and 9 are considered positively framed.

The SUS survey was followed by an interview with each participant to gather specific feedback and insights based on their artistic background and experience using the different tools. These processes provided valuable information on the strengths and weaknesses of each tool, as well as how they can be improved to better serve users. The following multiple-choice questions were also asked for evaluating the performance of each method:
\begin{enumerate}[label=\subscript{\textbf{Q}}{\textbf{{\arabic*}}}]
\setcounter{enumi}{10}
\item How much time did it take you to complete the image editing task using the tool you used in this study? [Much less time/About the same/Much more time]

\item How did you find each of the tools in terms of effectiveness in achieving the desired edits? [Very effective/Somewhat effective/Neutral/Somewhat ineffective/Very ineffective]

\item How does each of the tools you used perform in terms of time to complete the editing task? [Much faster/Somewhat faster/Acceptable/Somewhat slower/Much slower]

\item How likely are you to use each of these tools as an AI image editing tool in the future? [Very likely/Somewhat likely/Neutral/Somewhat unlikely/Very unlikely]

\end{enumerate}
\section{Findings}

\cref{fig:sample_inpainting} present the outcome of utilizing different fine-tuning techniques used in this work. For each method, all the images on columns 3-5 were generated by the participants during the user study. The original images showcased in column 1 were also generated by participants, following a consistent prompt and seed number provided to ensure uniformity across all users. Column 2, i.e., image with mask represents a potential region that can be selected for making edits using different methods. As demonstrated, Diffusion Brush is able to effectively make targeted adjustments to the image which are well-integrated with the image. In this section, we will also present the results of the user study in order to showcase the effectiveness of our method.
\subsection{Time measurement and the number of edits} As mentioned in section \ref{user-study}, we specified a maximum of 3 minutes for completing each of the tasks and also measured the average time spent to complete each fine-tuning task among different participants.  Despite the time constraint, users chose to continue using editing tools in an attempt to perform more fine-tuning. Therefore, we counted the number of edits (regardless of the quality) that participants were able to complete within 3 minutes using each tool. Based on this data, it was observed that the participants took the longest time while using Adobe Photoshop and on average they completed 1.2 edits in 3 minutes. On the other hand, the Diffusion Brush and Inpainting methods were relatively similar with an average of 5.1 and 5.9 edits in three minutes, respectively. This finding is further supported by the results of Q11, where users indicated that Photoshop takes much longer to complete an edit.

\subsection{Evaluation Survey results}
\begin{figure}[!ht]
  \centering
  
        \includegraphics[width=0.49\textwidth]{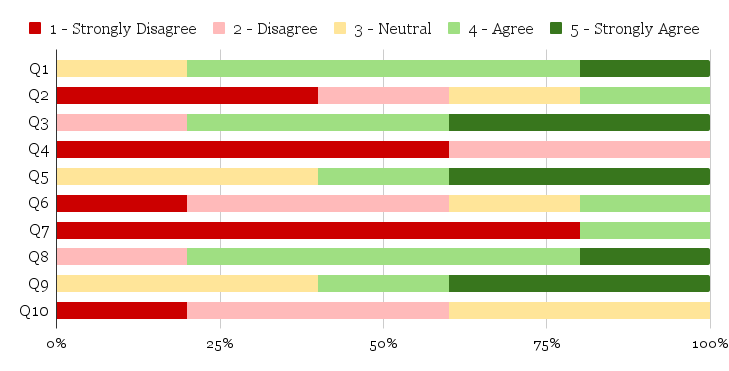}
\caption{Results of Q1 - Q10 for the usability of Diffusion Brush based on the tasks among different participants}
\label{dif-res}
\end{figure}

\begin{figure}[!ht]
  \centering
  
        \includegraphics[width=0.49\textwidth]{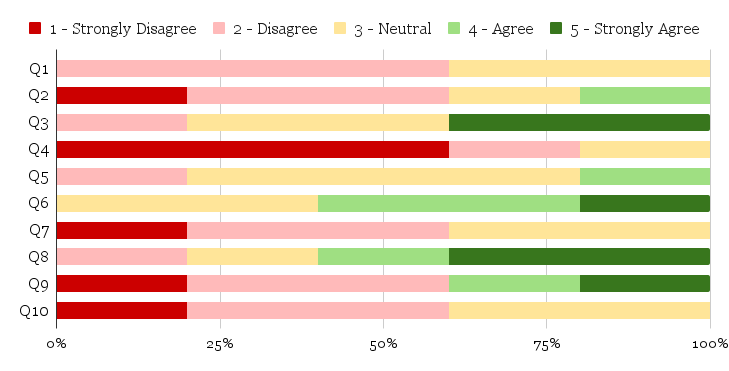}
\caption{Results of Q1 - Q10 for the usability of Inpainting based on the tasks among different participants}
\label{inp-res}
\end{figure}

\begin{figure}[!ht]
  \centering
  
        \includegraphics[width=0.49\textwidth]{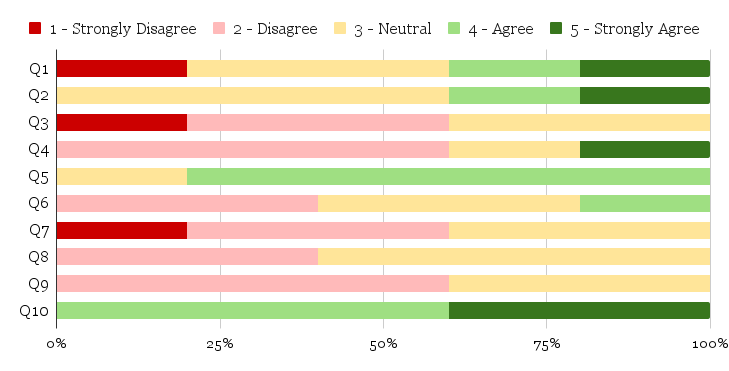}
\caption{Results of Q1 - Q10 for the usability of Photoshop based on the tasks among different participants}
\label{ps-res}
\end{figure}

Figures \ref{dif-res} \ref{inp-res}, and \ref{ps-res} present the results of the SUS survey among participants after using the Diffusion Brush, Inpainting, and Photoshop. For the positive phrasing questions, a higher score indicates higher usability, while for the negative phrasing questions, in contrast, a lower score indicates better usability.
Based on the bar charts, participants indicated that they are more likely to use Diffusion Brush compared to Photoshop and Inpainting, and that they find it the easiest tool to use. In addition, participants in Q4 expressed that they would not require technical assistance to use the system in the future, indicating its overall good design. These findings were further supported by the interview feedback. For example, when asked about their understanding of the different parameters in the tool, one participant stated: ``I believe that I understand the functionality of each parameter. I need to increase the mask strength value if I want to make bigger changes. The tool is quite intuitive and easy to use, and I think I can easily use it without needing any technical support." This feedback highlights that the tool has a user-friendly design and can be easily understood and used by a wide range of users. 

Results for the additional questions were also highly positive: users found Diffusion Brush to be faster and more effective than other techniques (Q12 means [Very Effective=5]: Diffusion Brush 4.2, Photoshop 3.75, Inpainting 2.8; Q13 means [Much faster=5]: Diffusion Brush 3.8, Photoshop 3.25, Inpainting 3.0), and expressed a strong likelihood to use the tool in the future (Q14 means [Very likely=5]: Diffusion Brush 4.6, Photoshop 3.25, Inpainting 3.0).

\subsection{Quantitative result}
Assessing a tool that integrates human edits presents a challenge due to the lack of a definitive reference. As a result, in our study, employing metrics like Fréchet Inception Distance (FID) or Learned Perceptual Image Patch Similarity (LPIPS) was unfeasible. Instead, we adopted ClipScore \cite{hessel2021clipscore} as our quantitative evaluation metric. This allowed us to gauge the image quality produced by each method and the alignment of each method's output with the actual content of the original prompt.
\cref{tab:results} presents the ClipScores for each method, along with the original generated image. As shown, on average, Diffusion Brush improves the overall ClipScore by 4.5\% compared to the original image, 6.66\% compared to inpainting and 3.59\% compared to the edited images of Photoshop.

We also present the SUS scores from the user study. Following the conventional methodology explained in \cite{brooke1996sus}, we obtained the SUS scores (on a 0-100 scale) and then averaged them over all participants. As shown in \cref{tab:results}, Diffusion Brush obtains ``Acceptable" performance on the acceptability scale \cite{bangor2009sus} (vs. ``Not acceptable" for Photoshop and ``Marginal" for Inpainting), and the highest SUS score among all methods tested.

\begin{table}
  \centering
  \begin{tabular}{@{}l|cc@{}}
    \toprule
     method & ClipScore $\uparrow$ & SUS score $\uparrow$ \\
    \midrule
    Original Image & 0.2924 & NA \\
    Photoshop & 0.2951 & 46 $\pm$ 15.268\\
    Inpainting & 0.2866 &58.5 $\pm$	17.818 \\ 
    Diffusion Brush & \textbf{0.3057} & \textbf{75} $\pm$	19.764\\
    \bottomrule
  \end{tabular}
  \caption{Quantitative results: We present the mean ClipScore $\pm$ std for the original images as well as all the images generated during the user study, corresponding to each of the methods. We also present the average SUS scores for each method across all participants.}
  \label{tab:results}
\end{table}

\section{Discussion and Future Work}
The Diffusion Brush offers the unique advantage of generating an unlimited number of small adjustments to an image. By setting a mask and fixing the hyperparameters, users can experiment with various random seed generator settings to create a vast array of localized changes. This unlimited potential for adjustment has been highly praised by artists who used the system during the user study. They highlighted this feature as one of the major strengths of the system and appreciated the ability to explore a variety of possibilities for their work. This sets the Diffusion Brush apart from traditional editing software that often has limited options for localized adjustments.

 The diffusion brush was purposefully developed to address a specific demographic who ultimately utilize the tool for the meticulous refinement of AI-generated images. An essential consideration revolves around ensuring that the Diffusion Brush's practicality aligns with the requirements of its intended user cohort. It is imperative to acknowledge that the efficacy of a tool like the Diffusion Brush remains constrained if it does not align with the usability expectations of its designated user base. The SUS results demonstrated that the Diffusion Brush is more usable than the other methods. Consequently, we highlight that one of the core contributions of the proposed method usability is the enhancement of usability.

Although the results of the SUS survey indicated that the participants had a good understanding of the functionality of the system controls, during the interviews, one of the concerns expressed by the participants was their ability to fully interpret the meaning of each parameter and understand its relationship with the other parameters. One participant commented, "I feel like these control parameters are correlated which makes it a bit challenging for me to control the system. I think once I have more time to discover and play with the system, I will be able to fully discover the relationship between these parameters." In response to this feedback, we decided to include a more detailed description of each parameter, its functionality, and its relationships in the methods section.

During the interviews, some participants suggested adding text-to-image editing capability to the Diffusion Brush. Three participants specifically mentioned this feedback. One participant stated, "I really like the tool as it is right now; it certainly provides value for me in my editing tasks and makes my life easier. But one feature that I would love to see is to be able to tell the system how to make these changes. I still want to use the masking editing, but if I can tell it what to do it would be great." We believe that this feature could provide value for making edits that are describable. However, as the participants stated, the current method can still be useful if users want to make adjustments that cannot be done using textual prompts. In the future, we are planning to combine diffusion brush with methods similar to null-text inversion \cite{mokady2023null} to enable editing of non-AI-generated images, as well as providing more flexibility in the fine-tuning process. 

\section{Conclusion}
In this paper, we presented Diffusion Brush, a novel LDM-based tool for fine-tuning AI-generated images. At its core, our method presents new random noise patterns at targeted regions in the reverse diffusion process and combines them with the original image latent intermediately, enabling the model to make targeted changes while keeping the rest of the image intact.  The formulation of the Diffusion Brush, combined with unique design choices, makes it a fast, reliable, and user-friendly tool that can be used by artists to make desired adjustments to the generated images. We also performed a thorough user study and demonstrated the effectiveness of Diffusion Brush in comparison to other widely used AI tools and image editing software. We believe that our work has important implications for the future of digital art fine-tuning and look forward to seeing how this tool is used in practice.

{\small
\bibliographystyle{ieee_fullname}
\bibliography{sample-base}
}


\end{document}